\documentclass{article} 
\usepackage[final]{neurips_2019}

\usepackage{amsmath,amsfonts,bm}

\def\eqref#1{equation~\ref{#1}}

\def\1{\bm{1}}

\DeclareMathAlphabet{\mathsfit}{\encodingdefault}{\sfdefault}{m}{sl}
\SetMathAlphabet{\mathsfit}{bold}{\encodingdefault}{\sfdefault}{bx}{n}

\newcommand{\sigmoid}{\sigma}

\usepackage{hyperref}
\usepackage{url}
\usepackage{wrapfig}
\usepackage{multirow}
\usepackage{graphicx}
\usepackage{soul}
\usepackage{enumitem}
\usepackage{hhline}
\title{Multi-domain Dialogue State Tracking as Dynamic Knowledge Graph Enhanced Question Answering}

\author{Li Zhou \\ Amazon Alexa Search \\lizhouml@amazon.com
\And
Kevin Small \\ Amazon Alexa Search \\ smakevin@amazon.com
}

\begin{document}
\maketitle

\begin{abstract}
Multi-domain dialogue state tracking (DST) is a critical component for conversational AI systems. The domain ontology (i.e., specification of domains, slots, and values) of a conversational AI system is generally incomplete, making the capability for DST models to generalize to new slots, values, and domains during inference imperative. In this paper, we propose to model multi-domain DST as a question answering problem, referred to as {\em Dialogue State Tracking via Question Answering} (DSTQA). Within DSTQA, each turn generates a question asking for the value of a (domain, slot) pair, thus making it naturally extensible to unseen domains, slots, and values. Additionally, we use a dynamically-evolving knowledge graph to explicitly learn relationships between (domain, slot) pairs. Our model has a $5.80\%$ and $12.21\%$ relative improvement over the current state-of-the-art model on MultiWOZ 2.0 and MultiWOZ 2.1 datasets, respectively. Additionally, our model consistently outperforms the state-of-the-art model in domain adaptation settings.
\end{abstract}

\section{Introduction}
\begin{figure}
  \includegraphics[width=1.0\linewidth]{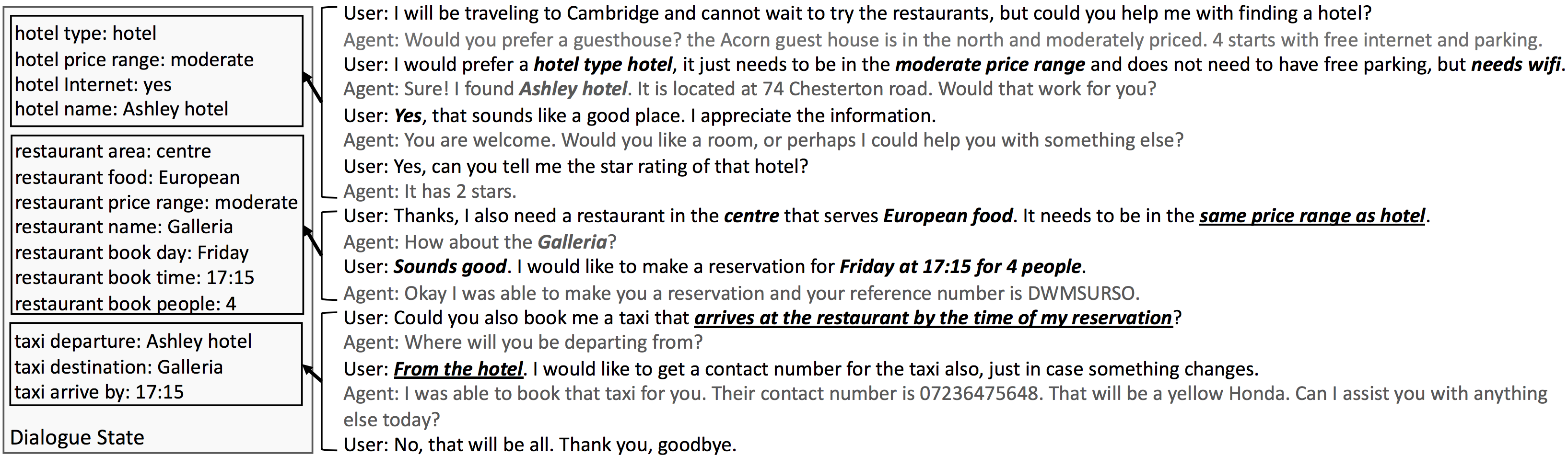}
  \caption{The right-hand side shows a dialogue which involves $3$ domains, and the left-hand side shows its dialogue state in the end. Bold text indicate mentions and paraphrases of slot values. Underlined text indicates scenarios where multi-turn inference is required.}
  \label{fig:dialogue_example}
  \vspace{-0.5cm}
\end{figure}

In a task-oriented dialogue system, the dialogue policy determines the next action to perform and next utterance to say based on the current dialogue state. A dialogue state defined by {\em frame-and-slot semantics} is a set of (key, value) pairs specified by the domain ontology~\citep{Jurafsky2000}. A key is a (domain, slot) pair and a value is a slot value provided by the user. Figure \ref{fig:dialogue_example} shows a dialogue and state in three domain contexts.
Dialogue state tracking (DST) in multiple domains is a challenging problem. First of all, in production environments, the domain ontology is being continuously updated such that the model must generalize to new values, new slots, or even new domains during inference. Second, the number of slots and values in the training data are usually quite large. For example, the MultiWOZ $2.0/2.1$ datasets~\citep{budzianowski2018multiwoz,eric2019multiwoz} have $30$ (domain, slot) pairs and more than $4,500$ values~\citep{wu-etal-2019-transferable}. As the model must understand slot and value paraphrases, it is infeasible to train each slot or value independently. Third, multi-turn inferences are often required as shown in the underlined areas of Figure \ref{fig:dialogue_example}.

Many single-domain DST algorithms have been proposed~\citep{mrkvsic2017neural,ren2018towards,zhong2018global}.
For example,
~\cite{zhong2018global} learns a local model for each slot and a global model shared by all slots.
However, single domain models are difficult to scale to multi-domain settings, leading to the development of multi-domain DST algorithms. For example,~\citet{gce} improves~\citet{zhong2018global}'s work by removing local models and building a slot-conditioned global model to share parameters between domains and slots,
thus computing a score for every (domain, slot, value) tuple. This approach remains problematic for settings with a large value set (e.g., {\em user phone number}). ~\citet{wu-etal-2019-transferable} proposes an encoder-decoder architecture which takes dialogue contexts as source sentences and state annotations as target sentences, but does not explicitly use relationships between domains and slots. For example, if a user booked a restaurant and asks for a taxi, then the destination of the taxi is likely to be that restaurant, and if a user booked a $5$ star hotel, then the user is likely looking for an expensive rather than a cheap restaurant. As we will show later, such relationships between domains and slots help improve model performance.

To tackle these challenges, we propose DSTQA ({\em Dialogue State Tracking via Question Answering)}, a new multi-domain DST model inspired by recently developed reading comprehension and question answering models. Our model reads dialogue contexts to answer a series of questions that asks for the value of a (domain, slot) pair. 
Specifically, we construct two types of questions: 1) multiple choice questions for (domain, slot) pairs with a limited number of value options and 2) span prediction questions, of which the answers are spans in the contexts, designed for (domain, slot) pairs that have a large or infinite number of value options. Finally,
we represent (domain, slot) pairs as a dynamically-evolving knowledge graph with respect to the dialogue context, and utilize this graph to drive improved model performance. Our contributions are as follows: (1) we propose to model multi-domain DST as a question answering problem such that tracking new domains, new slots and new values is simply constructing new questions, (2) we propose using a bidirectional attention~\citep{seo2016bidirectional} based model for multi-domain dialogue state tracking, and (3) we extend our algorithm with a dynamically-evolving knowledge graph to further exploit the structure between domains and slots.

\label{sec:model}

\section{Problem Formulation}
In a multi-domain dialogue state tracking problem, there are $M$ domains $D=\{d_1, d_2, ..., d_M\}$. For example, in MultiWOZ 2.0/2.1 datasets, there are 7 domains: {\em restaurant}, {\em hotel}, {\em train}, {\em attraction}, {\em taxi}, {\em hospital}, and {\em police}. Each domain $d \in D$ has $N^d$ slots $S^d = \{s^d_1, s^d_2, ...,s^d_{N^d}\}$, and each slot $s \in S^d$ has $K^s$ possible values $V^s=\{v^s_1, v^s_2, ...,v^s_{K^s}\}$. For example, the {\em restaurant} domain has a slot named {\em price range}, and the possible values are {\tt cheap}, {\tt moderate}, and {\tt expensive}. Some slots do not have pre-defined values, that is, $V^s$ is missing in the domain ontology. For example, the {\em taxi} domain has a slot named {\em leave time}, but it is a poor choice to enumerate all the possible leave times the user may request as the size of $V^s$ will be very large. Meanwhile, the domain ontology can also change over time. Formally, we represent a dialogue $X$ as $X=\{U^a_1, U^u_1, U^a_2, U^u_2, ..., U^a_T, U^u_T\}$, where $U^a_t$ is the agent utterance in turn $t$ and $U^u_t$ is the user utterance in turn $t$. Each turn $t$ is associated with a dialogue state $\text{y}_t$. A dialogue state $\text{y}_t$ is a set of (domain, slot, value) tuples. Each tuple represents that, up to the current turn $t$, a slot $s \in S^d$ of domain $d \in D$, which takes the value $v \in V^s$ has been provided by the user. Accordingly, $\text{y}_t$'s are targets that the model needs to predict.

\section{Multi-domain Dialogue State Tracking via Question Answering (DSTQA)}
\label{sec:dstqa}
We model multi-domain DST as a question answering problem and use machine reading methods to provide answers. To predict the dialogue state at turn $t$, the model observes the context $C_t$, which is the concatenation of $\{U_1^a, U_1^u, ..., U_t^a, U_t^u\}$. The context is read by the model to answer the questions defined as follows.
First, for each domain $d \in D$ and each slot $s \in S^d$ where there exists a pre-defined value set $V^s$, we construct a question $Q_{d,s} = \{d, s, V^s, \text{{\tt not mentioned}}, \text{{\tt don't care}}\}$. That is, a question is a set of words or phrases which includes a domain name, a slot name, a list of all possible values, and two special values {\tt not mentioned} and {\tt don't care}. One example of the constructed question for {\em restaurant} domain and {\em price range} slot is $Q_{d,s} = \{ \text{{\em restaurant}}, \text{{\em price range}}, \text{{\tt cheap}}, \text{{\tt moderate}}, \text{{\tt expensive}}, \text{{\tt not mentioned}}, \text{{\tt don't care}} \}$. The constructed question represents the following natural language question:

{\em
``In the dialogue up to turn $t$, did the user mention the `price range' of the `restaurant' he/she is looking for? If so, which of the following option is correct: A) {\tt cheap}, B) {\tt moderate}, C) {\tt expensive}, D) {\tt don't care}.''
}

As we can see from the above example, instead of only using domains and slots to construct questions (corresponding to natural language questions {\em what is the value of this slot?}), we also add candidate values $V^s$ into $Q_{d,s}$, this is because values can be viewed as descriptions or complimentary information to domains and slots. For example, \text{{\tt cheap}}, \text{{\tt moderate}} and \text{{\tt expensive}} explains what {\em price range} is. In this way, the constructed question $Q_{d,s}$ contains rich information about the domains and slots to predict, and easy to generalize to new values.

In the case that $V^s$ is not available, the question is just the domain and slot names along with the special values, that is, $Q_{d,s} = \{d, s, \text{{\tt not mentioned}}, \text{{\tt don't care}}\}$. For example, the constructed question for {\em train} domain and {\em leave time} slot is $Q_{d,s} = \{\text{{\em train}}, \text{{\em leave time}}, \text{{\tt not mentioned}}, \text{{\tt don't care}}\}$, and represents the following natural language question:

{\em 
``In the dialogue up to turn $t$, did the user mention the `leave time' of the `train' he/she is looking for? If so, what is the `leave time' the user preferred?''
}

The most important concept to note here is that the proposed DSTQA model can be easily extended to new domains, slots, and values. Tracking new domains and slots is simply constructing new queries, and tracking new values is simply extending the constructed question of an existing slot.

Although we formulate multi-domain dialogue state tracking as a question answering problem, we want to emphasize that there are some fundamental differences between these two settings. In a standard question answering problem, question understanding is a major challenge and the questions are highly dependent on the context where questions are often of many different forms~\citep{rajpurkar2018know}. Meanwhile, in our formulation, the question forms are limited to two, every turn results in asking a restricted set of question types, and thus question understanding is straightforward. Conversely, our formulation has its own complicating characteristics including: (1) questions in consecutive turns tend to have the same answers, (2) an answer is either a span of the context or a value from a value set, and (3) the questions we constructed have some underlying connections defined by a dynamically-evolving knowledge graph (described in Section \ref{sec:graph}), which can help improve model performance.
In any case, modeling multi-domain DST with this approach allows us to easily transfer knowledge to new domains, slots, and values simply by constructing new questions.
Accordingly, many existing reading comprehension algorithms~\citep{seo2016bidirectional,yu2018qanet,devlin2019bert,clark2018simple} can be directly applied here. In this paper, we propose a bidirectional attention flow~\citep{seo2016bidirectional} based model for multi-domain DST.
\subsection{Model Overview}
\label{sec:model}
\begin{figure}[htb]
  \includegraphics[width=1\linewidth]{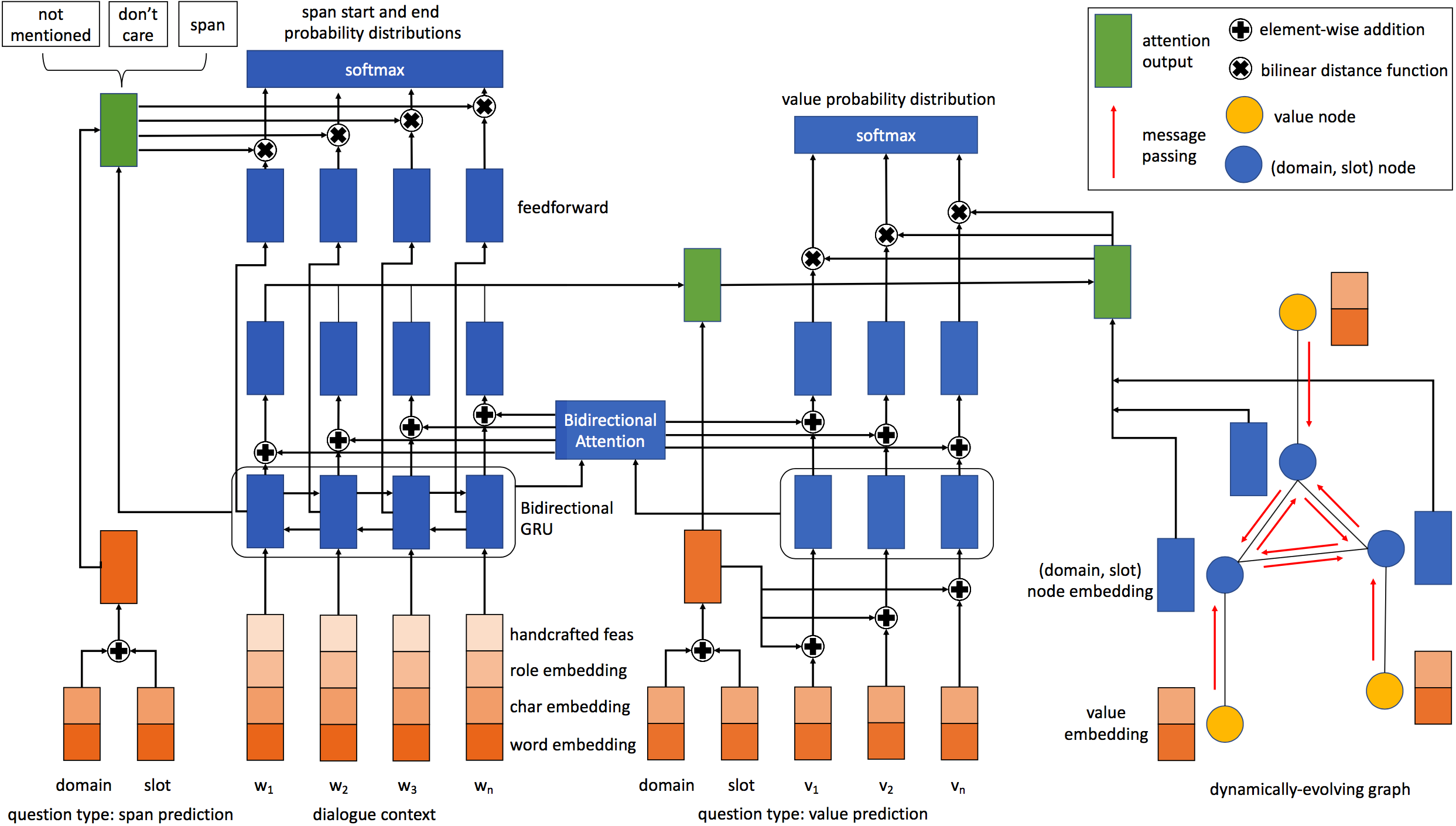}
  \caption{DSTQA model architecture. When the question type is value prediction, a bidirectional attention layer is applied to the dialogue context and the question, and a graph embedding is injected to the output of the bidirectional attention layer. When the question type is span prediction, the question is used to attend over the dialogue context to predict span start and end positions.}
  \label{fig:model}
  \vspace{-0.3cm}
\end{figure}

Figure \ref{fig:model} summarizes the DSTQA architecture, where notable subcomponents are detailed below.
%
\textbf{1.\ Word Embedding Layer}: 
For each word in context $C_t$, similar to \cite{seo2016bidirectional}, we apply a character embedding layer based on convolutional neural network to get a $D^{\text{Char}}$ dimensional character-level embedding. We then adopt ELMo~\citep{peters2018deep}, a deep contextualized word representations, to get a $D^{\text{ELMo}}$ dimensional word-level embedding. Other contextualized word embeddings such as BERT~\citep{devlin2019bert} can also be applied here but is orthogonal to DSTQA and is left for future work. The final word embedding of context $C_t$ is the concatenation of the character-level embedding and the ELMo embedding, and is denoted by $W^c \in \mathbb{R}^{L_c \times D^w}$, where $L_c$ is the number of words in context $C_t$ and $D^w = D^{\text{ELMo}} + D^{\text{Char}}$.
Similarly, For a question $Q_{d, s}$, we treat each element in $Q_{d,s}$ (either a domain name, a slot name, or a value from the value set) as a sentence and compute its word embedding.
We then take the mean of the word embeddings in each element as the embedding of that element. Then the question embedding is represented by a set $\{w^{d} \in \mathbb{R}^{D^w}, w^{s} \in \mathbb{R}^{D^w}, W^{\bar{v}} \in \mathbb{R}^{L_{\bar{v}} \times D^w}\}$, where $w^d$, $w^s$ and $W^{\bar{v}}$ are domain, slot and value embeddings, respectively, and $L_{\bar{v}}$ is the number of values in $V^s$ plus {\tt not mentioned} and {\tt don't care}. To represent the question embedding as one single matrix, we define $W^q \in \mathbb{R}^{L_{\bar{v}} \times D^w}$, where each row of $W^q$ is calculauted by $W^q_{j,:} = w^d + w^s + W^{\bar{v}}_{j,:}$.

\textbf{2.\ Context Encoding Layer}: 
We apply a bidirectional GRU to encode the context $C_t$. Denoting the $i$-th word in the context $C_t$ by $w_i$, then the input to the bidirectional GRU at time step $i$ is the concatenation of the following three vectors: 1) $w_i$'s word embeddings, $W^c_{i, :}$, 2) the corresponding role embedding, and 3) exact match features. There are two role embeddings: the agent role embedding $e_a \in \mathbb{R}^r$ and the user role embedding $e_u \in \mathbb{R}^r$ where both are trainable. Exact match features are binary indicator features where for each (domain, slot) pair, we search for occurrences of its values in the context in original and lemmatized forms. Then for each (domain, slot) pair, we use two binary features to indicate whether $w_i$ belongs to an occurrence in either form.
The final output of this layer is a matrix $E^c \in \mathbb{R}^{L_c \times D^\text{biGRU}}$, where $L_c$ is the number of words in the context $C_t$ and $D^\text{biGRU}$ is the dimension of bidirectional GRU's hidden states (includes both forward and backward hidden states). In our experiments, we set $D^\text{biGRU}$ equals to $D^w$.

\textbf{3.\ Question-Context Bidirectional Attention Layer}: 
Inspired by~\citet{seo2016bidirectional}, we apply a bidirectional attention layer which computes attention in two directions: from context $C_t$ to question $Q_{d,s}$, and from question $Q_{d,s}$ to context $C_t$.
To do so, we first define an attention function $\mathbb{R}^{m*n} \times \mathbb{R}^n \rightarrow \mathbb{R}^m$ that will be used frequently in the following sections. The inputs to the function are a key matrix $K \in \mathbb{R}^{m * n}$ and a query vector $q \in \mathbb{R}^{n}$. The function calculates the attention score of $q$ over each row of $K$. Let $O \in \mathbb{R}^{m*n}$ be a matrix which is $q$ repeated by $m$ times, that is, $O_{:,j} = q$ for all $j$. Then, the attention function is defined as:
\begin{gather*}
\text{Att}_{\beta}(K, q) = \text{Softmax}([K; O^\top; K \odot O^\top] \cdot \beta)
\end{gather*}
Where $\beta \in \mathbb{R}^{3 n}$ are learned model parameters, $\odot$ is the element-wise multiplication operator, and $[;]$ is matrix row concatenation operator. We use subscript of $\beta$, $\beta_i$, to indicate different instantiations of the attention function.

The attention score of a context word $w_i$ to values in $Q_{d,s}$ is given by $\alpha^{v}_i = \text{Att}_{\beta_1}(W^q, E^c_{i,:}) \in \mathbb{R}^{L_{\bar{v}}}$, and the attention score of a value $v_j$ to context words in $C_t$ is given by $\alpha^{w}_j = \text{Att}_{\beta_1}(E^c, W^q_{j, :}) \in \mathbb{R}^{L_{c}}$. $\beta_1$ is shared between these two attention functions.
Then, the question-dependent embedding of context word $w_i$  is $B^{QD}_i = {W^q}^\top \cdot \alpha^{v}_i$ and can be viewed as the representation of $w_i$ in the vector space defined by the question $Q_{d,s}$. Similarly, the context-dependent embedding for value $v_j$ is $B^{CD}_j = {E^c}^\top \cdot \alpha^{w}_j$ and can be viewed as the representation of $v_j$ in the vector space defined by the context $C_t$. The final context embedding is $B^c = E^c + B^{QD} \in \mathbb{R}^{L_c \times D^w}$ and the final question embedding is $B^q = B^{CD} + W^q \in \mathbb{R}^{L_{\bar{v}} \times D^w}$. 

\textbf{4.\ Value Prediction Layer}:
When $V^s$ exists in $Q_{d,s}$, we calculate a score for each value in $Q_{d,s}$, and select the one with the highest score as the answer. First, we define a bilinear function $\mathbb{R}^{m*n} \times \mathbb{R}^n \rightarrow \mathbb{R}^m$. It takes a matrix $X \in \mathbb{R}^{m*n}$ and a vector $y \in \mathbb{R}^n$, returning a vector of length $m$,
\begin{align*}
\text{BiLinear}_{\Phi}(X, y) = X^\top \Phi y
\end{align*}
where $\Phi \in \mathbb{R}^{n*n}$ are learned model parameters. Again, we use subscript of $\Phi$, $\Phi_i$, to indicate different instantiations of the function.

We summarize context $B^c$ into a single vector with respect to the domain and slot and then apply a bilinear function to calculate the score of each value. More specifically,
We calculate the score of each value $v$ at turn $t$ by
\begin{align}
p^v_t = \text{Softmax}\left(\text{BiLinear}_{\Phi_1}\left(B^q, {B^c}^\top \cdot \alpha^b\right)\right)
\label{eq:vscore}
\end{align}
where $\alpha^b = \text{Att}_{\beta_2}(B^c, w^d + w^s) \in \mathbb{R}^{L_c}$ is the attention score over $B^c$, and $p^v_t \in \mathbb{R}^{L_{\bar{v}}}$.
We calculate the cross entropy loss of the predicted scores by
$\text{Loss}_v = \sum_t \sum_{d \in D,s \in \hat{S}^d}\text{CrossEntropy}\left(p_t^v, y_t^v\right)$
where $y^v_t \in \mathbb{R}^{L_{\bar{v}}}$ is the label, which is the one-hot encoding of the true value of domain $d$ and slot $s$, and $\hat{S}^d$ is the set of slots in domain $d$ that has pre-defined $V^s$. 

\textbf{5.\ Span Prediction Layer}:
When the value set $V^s$ is unknown or too large to enumerate, such as {\em pick up time} in {\em taxi} domain, we predict the answer to a question $Q_{d,s}$ as either a span in the context or two special types: {\tt not mentioned} and {\tt don't care}. The span prediction layer has two components. The first component predicts the answer type of $Q_{d,s}$. The type of the answer is either {\tt not mentioned}, {\tt don't care} or {\tt span}, and is calculated by
$
p^{st}_t = \text{Softmax}(\Theta_1 \cdot (w^d + w^s + {E^c}^\top \cdot \alpha^e ) )
$
where $\alpha^e = \text{Att}_{\beta_3}(E^c, w^d + w^s) \in \mathbb{R}^{L_c}$, $\Theta_1 \in \mathbb{R}^{3 * D^w}$ is a model parameter to learn, and $p^{st}_t \in \mathbb{R}^3$. The loss of span type prediction is
$
\text{Loss}_{st} = \sum_t \sum_{d \in D, s\in \bar{S}^d} \text{CrossEntropy}\left(p^{st}_t, y^{st}_t\right)
$
where $y^{st}_t \in \mathbb{R}^3$ is the one-hot encoding of the true span type label, and $\bar{S}^d$ is the set of slots in domain $d$ that has no pre-defined $V^s$.
The second component predicts a span in the context corresponding to the answer of $Q_{d,s}$. To get the probability distribution of a span's start index, we apply a bilinear function between contexts and (domain, slot) pairs. More specifically,
\begin{align*}
p^{ss}_t =  \text{Softmax}\left(\text{BiLinear}_{\Phi_2}\left( \text{Relu}\left(E^c  \cdot \Theta_2\right), \left(w^d + w^s + {E^c}^\top \cdot \alpha^e \right) \right)\right)
\end{align*}
where $\Theta_2 \in \mathbb{R}^{D^w * D^w}$ and $p_t^{ss} \in \mathbb{R}^{L_c}$. The $\text{Bilinear}$ function's first argument is a non-linear transformation of the context embedding, and its second argument is a context-dependent (domain, slot) pair embedding. 
Similarly, the probability distribution of a span's end index is
$$
p^{se}_t = \text{Softmax}\left(\text{BiLinear}_{\Phi_3}\left( \text{Relu}\left(E^c \cdot \Theta_2 \cdot \Theta_3\right), \left(w^d + w^s + {E^c}^\top \cdot \alpha^e \right) \right) \right)
$$
where $\Theta_3 \in \mathbb{R}^{D^w * D^w}$ and $p_t^{se} \in \mathbb{R}^{L_c}$. 
The prediction loss is
$\text{Loss}_{span} = \sum_t \sum_{d \in D, s\in \bar{S}^d} 
\text{CrossEntropy}(p^{ss}_t, y^{ss}_t) + \text{CrossEntropy}(p^{se}_t, y^{se}_t)$
where $y^{ss}_t, y^{se}_t \in \mathbb{R}^{L_c}$ is one-hot encodings of true start and end indices, respectively.
The score of a span is the multiplication of probabilities of its start and end index.
The final loss function is:
$
\text{Loss} = \text{Loss}_v + \text{Loss}_{st} + \text{Loss}_{span} 
$.
In most publicly available dialogue state tracking datasets, span start and end labels do not exist. In Section \ref{sec:impl_detail} we will show how we construct these labels. 
\section{Dynamic Knowledge Graph for Multi-domain dialogue State Tracking}
\label{sec:graph}
In our problem formulation, at each turn, our proposed algorithm asks a set of questions, one for each (domain, slot) pair. In fact, the (domain, slot) pairs are not independent. For example, if a user requested a train for $3$ people, then the number of people for hotel reservation may also be $3$. If a user booked a restaurant, then the destination of the taxi is likely to be that restaurant.  Specifically, we observe four types of relationships between (domain, slot) pairs in MultiWOZ $2.0$/$2.1$ dataset:
\begin{enumerate}
   \item $(s, r_v, s')$: a slot $s \in S^d$ and another slot $s' \in S^{d'}$ have the same set of possible values. That is, $V^s$ equals to $V^{s'}$. For example, in MultiWOZ $2.0$/$2.1$ dataset, domain-slot pairs ({\em restaurant}, {\em book day}) and ({\em hotel}, {\em book day}) have this relationship.
    \item $(s, r_s, s')$: the value set of a slot $s \in S^d$ is a subset of the value set of $s' \in S^{d'}$. For example, in MultiWOZ $2.0$/$2.1$ dataset, value sets of ({\em restaurant}, {\em name}), ({\em hotel}, {\em name}), ({\em train}, {\em station}) and ({\em attraction}, {\em name}) are subsets of the value set of ({\em taxi}, {\em destination}). 
    \item $(s, r_c, s')$: the informed value $v \in V^s$ of slot $s$ is correlated with the informed value $v \in V^{s'}$ of slot $s'$ even though $V^s$ and $V^{s'}$ do not overlap. For example, in MultiWOZ $2.0$/$2.1$ dataset, the price range of a reserved restaurant is correlated with the star of the booked hotel. This relationship is not explicitly given in the ontology.
    \item $(s, r_i, v)$: the user has informed value $v \in V^s$ of slot $s \in S^d$.
\end{enumerate}
In this section, we propose using a dynamic knowledge graph to further improve model performance by exploiting this information. We represent (domain, slot) pairs and values as nodes in a graph linked by the relationship defined above, and then propagate information between them. The graph is \emph{dynamically evolving}, since the fourth relationship above, $r_i$, depends on the dialogue context.

\subsection{Graph Definition}

The right-hand side of Figure \ref{fig:model} is an example of the graph we defined based on the ontology. There are two types of nodes $\{M, N\}$ in the graph. One is a (domain, slot) pair node representing a (domain, slot) pair in the ontology and another is a value node representing a value from a value set. For a domain $d \in D$ and a slot $s \in S^d$, we denote the corresponding node by $M_{d,s}$, and for a value $v \in V^s$, we denote the corresponding node by $N_{v}$.
There are also two types of edges. One type is the links between $M$ and $N$. At each turn $t$, if the answer to question $Q_{d, s}$ is $v \in V^s$, then $N_v$ is added to the graph and linked to $M_{d,s}$. By default, $M_{d, s}$ is linked to a special {\tt not mentioned} node.
The other type of edges is links between nodes in $M$. Ideally we want to link nodes in $M$ based on the first three relationships described above. However, while $r_v$ and $r_s$ are known given the ontology, $r_c$ is unknown and cannot be inferred just based on the ontology. As a result, we connect every node in $M$ (i.e. the (domain, slot) pair nodes) with each other, and let the model to learn their relationships with an attention mechanism, which will be described shortly.

\subsection{Attention Over the Graph}
We use an attention mechanism to calculate the importance of a node's neighbors to that node, and then aggregate node embeddings based on attention scores. 
~\citet{velivckovic2018graph} describes a graph attention network, which performs self-attention over nodes. In contrast with their work, we use dialogue contexts to attend over nodes.

Our attention mechanism has two steps. The first step is to propagate the embedding of $N_v$ to its linked $M_{d,s}$, so that the embedding of $M_{d,s}$ depends on the value prediction from previous turns. We propagate $N_v$'s embedding by
$g_{d,s} = \eta (w^d + w^s) + (1-\eta) \sigmoid(\Theta_4 \cdot W_{v,:}^{\bar{v}})$
where $g_{d,s} \in \mathbb{R}^{D^w}$ is the new embedding of $M_{d,s}$, $\eta \in [0, 1]$ is a hyper-parameter, and $\Theta_4 \in \mathbb{R}^{D^w \times D^w}$ is a model parameter to learn. $g_{d,s}$ essentially carries the following information: in previous turns, the user has mentioned value $v$ of a slot $s$ from a domain $d$.
In practice, we find out that simply adding $w^d$, $w^s$ and $W^{\bar{v}}$ yields the best result. That is $g_{d,s} = w^d + w^s + W_{v,:}^{\bar{v}}$.
The second step is to propagate information between nodes in $M$. 
For each domain $d$ and slot $s$, ${B^c}^\top \cdot \alpha^b$ in Equation (\ref{eq:vscore}) is the summarized context embedding with respect to $d$ and $s$. We use this vector to attend over all nodes in $M$, and the attention score is $\alpha^g = \text{Att}_{\beta_4}(G, {B^c}^\top \cdot \alpha^b)$, where $G \in \mathbb{R}^{|M| * D^w}$ is a matrix stacked by $g_{d,s}^\top$. The attention scores can be interpreted as the learned relationships between the current (domain, slot) node and all other (domain, slot) nodes.
Using context embeddings to attend over the graph allows the model to assign attention score of each node based on dialogue contexts. Finally,
%
The graph embedding is $z_{d,s} = G \cdot \alpha^g$.
We inject $z_{d,s}$ to Equation ($\ref{eq:vscore}$) with a gating mechanism:
\begin{gather}
p^v_t = \text{Softmax}\left(\text{BiLinear}_{\Phi_1}\left(B^q, (1-\gamma) {B^c}^\top \cdot \alpha^b + \gamma z_{d,s}\right)\right) 
\label{eq:graph}
\end{gather}
where $\gamma=\sigmoid({B^c}^\top \cdot \alpha^b + z_{d,s})$ is the gate and controls how much graph information should flow to the context embedding given the dialogue context. Some utterances such as {\em ``book a taxi to Cambridge station"} do not need information in the graph, while some utterances such as {\em ``book a taxi from the hotel to the restaurant"} needs information from other domains. $\gamma$ dynamically controls in what degree the graph embedding is used.
and graph parameters are trained together with all other parameters.

\section{Experiments}
We evaluate our model on three publicly available datasets: (non-multi-domain) WOZ $2.0$~\citep{mrkvsic2017neural}, MultiWOZ $2.0$~\citep{budzianowski2018multiwoz}, and MultiWOZ $2.1$~\citep{eric2019multiwoz}. Due to limited space, please refer to Appendix \ref{ap:woz2} for results on (non-multi-domain) WOZ $2.0$ dataset.
MultiWOZ $2.0$ dataset is collected from a Wizard of Oz style experiment and has $7$ domains: {\em restaurant}, {\em hotel}, {\em train}, {\em attraction}, {\em taxi}, {\em hospital}, and {\em police}. Similar to \citet{wu-etal-2019-transferable}, we ignore the {\em hospital} and {\em police} domains because they only appear in training set. There are $30$ (domain, slot) pairs and a total of $10438$ task-oriented dialogues. A dialogue may span across multiple domains. For example, during the conversation, a user may book a restaurant first, and then book a taxi to that restaurant. For both datasets, we use the train/test splits provided by the dataset. The domain ontology of the datasets is described in Appendix \ref{ap:ontology}. MultiWOZ $2.1$ contains the same dialogues and ontology as MultiWOZ $2.0$, but fixes some annotation errors in MultiWOZ $2.0$.

Two common metrics to evaluate dialogue state tracking performance are \textbf{\emph{Joint}} accruacy and \textbf{\emph{Slot}} accuracy. Joint accuracy is the accuracy of dialogue states. A dialogue state is correctly predicted only if all the values of (domain, slot) pairs are correctly predicted. Slot accuracy is the accuracy of (domain, slot, value) tuples. A tuple is correctly predicted only if the value of the (domain, slot) pair is correctly predicted. In most literature, joint accuracy is considered as a more challenging and more important metric. 

\subsection{Implementation Details}
\label{sec:impl_detail}
Existing dialogue state tracking datasets, such as MultiWOZ $2.0$ and MultiWOZ 2.1, do not have annotated span labels but only have annotated value labels for slots. As a result, we preprocess MultiWOZ $2.0$ and MultiWOZ $2.1$ dataset to convert value labels to span labels: we take a value label in the annotation, and search for its last occurrence in the dialogue context, and use that occurrence as span start and end labels.
There are $30$ slots in MultiWOZ $2.0$/$2.1$ dataset, and $5$ of them are time related slots such as {\em restaurant book time} and {\em train arrive by}, and the values are 24-hour clock time such as {\tt 08:15}. We do span prediction for these $5$ slots and do value prediction for the rest of slots because it is not practical to enumerate all time values. We can also do span prediction for other slots such as {\em restaurant name} and {\em hotel name} with the benefit of handling out-of-vocabulary values, but we leave these experiments as future work.
WOZ $2.0$ dataset only has one domain and $3$ slots, and we do value prediction for all these slots without graph embeddings.

We implement our model using AllenNLP~\citep{Gardner2017AllenNLP} framework.\footnote{https://github.com/alexa/dstqa} For experiments with ELMo embeddings, we use a pre-trained ELMo model\footnote{https://allennlp.org/elmo} in which the output size is $D^{ELMo} = 512$. The dimension of character-level embeddings is $D^{Char} = 100$, making $D^w = 612$. ELMo embeddings are fixed during training. For experiments with GloVe embeddings, we use GloVe embeddings pre-trained on Common Crawl dataset.\footnote{https://nlp.stanford.edu/projects/glove/} The dimension of GloVe embeddings is $300$, and the dimension of character-level embeddings is 100, such that $D^w = 400$. GloVe embeddings are trainable during training. The size of the role embedding is $128$. The dropout rate is set to $0.5$. We use Adam as the optimizer and the learning rate is set to $0.001$. We also apply word dropout that randomly drop out words in dialogue context with probability $0.1$.

When training DSTQA with the dynamic knowledge graph, in order to predict the dialogue state and calculate the loss at turn $t$, we use the model with current parameters to predict the dialogue state up until turn $t-1$, and dynamically construct a graph for turn $t$. We have also tried to do teacher forcing which constructs the graph with ground truth labels (or sample ground truth labels with an annealed probability), but we observe a negative impact on joint accuracy. On the other hand, target network~\citep{mnih2015human} may be useful here and will be investigated in the future. More specifically, we can have a copy of the model that update periodically, and use this model copy to predict dialogue state up until turn $t-1$ and construct the graph.

\subsection{Results on MultiWoz 2.0 and MultiWOZ 2.1 dataset.}
\begin{wraptable}{r}{0pt}
\resizebox{0.35\textwidth}{!}{%
\begin{tabular}{|l|c|c|}
\hline
                                                       & Joint  & Slot  \\ \hline
GLAD                                                   & 35.57          & 95.44         \\ \hline
GCE                                                    & 36.27          & 98.42         \\ \hline
Neural Reading                                                    & 42.12         & -        \\ \hline
SUMBT                                                  & 46.65         & 96.44         \\ \hline
TRADE                                                  & 48.62          & 96.92         \\ \hhline{|=|=|=|}
DSTQA w/span    & \textbf{51.36}  &  97.22    \\ \hline
-graph  &   50.89 &     97.17 \\ \hline
-gating &   50.38  &   97.14  \\ \hline
-bi att +avg &  49.74  &   97.11 \\ \hline
-bi att   & 49.51          & 97.07         \\ \hline
-ELMo +GloVe   & 49.52          & 96.96         \\ \hhline{|=|=|=|}
DSTQA w/o span   & \textbf{51.44}          & 97.24         \\ \hline 
-ELMo +GloVe   & 50.81          & 97.19         \\ \hline

\end{tabular}%
}
\caption{Results on MultiWOZ $2.0$ dataset.}
\label{exp:alldomain}
\end{wraptable}
We first evaluate our model on MultiWOZ 2.0 dataset as shown in Table \ref{exp:alldomain}. We compare with five published baselines. TRADE~\citep{wu-etal-2019-transferable} is the current published state-of-the-art model. It utilizes an encoder-decoder architecture that takes dialogue contexts as source sentences, and takes state annotations as target sentences. SUMBT~\citep{lee2019sumbt} fine-tunes a pre-trained BERT model~\citep{devlin2019bert} to learn slot and utterance representations. Neural Reading~\citep{gao2019dialog} learns a question embedding for each slot, and predicts the span of each slot value. GCE~\citep{gce} is a model improved over GLAD~\citep{zhong2018global} by using a slot-conditioned global module. Details about baselines are in Section \ref{sec:related_works}. 

For our model, we report results under two settings. In the DSTQA w/span setting, we do span prediction for the five time related slots as mentioned in Section \ref{sec:impl_detail}. This is the most realistic setting as enumerating all possible time values is not practical in a production environment.  In the DSTQA w/o span setting, we do value prediction for all slots, including the five time related slots. To do this, we collect all time values appeared in the training data to create a value list for time related slots as is done in baseline models. It works in these two datasets because there are only $173$ time values in the training data, and only $14$ out-of-vocabulary time values in the test data. Note that in all our baselines, values appeared in the training data are either added to the vocabulary or added to the domain ontology, so DSTQA w/o span is still a fair comparison with the baseline methods.
Our model outperforms all models. DSTQA w/span has a $5.64\%$ relative improvement and a $2.74\%$ absolute improvement over TRADE. We also show the performance on each single domain in Appendix \ref{ap:singledomain}. DSTQA w/o span has a $5.80\%$ relative improvement and a $2.82\%$ absolute improvement over TRADE.
We can see that DSTQA w/o span performs better than DSTQA w/span, this is mainly because we introduce noises when constructing the span labels, meanwhile, span prediction cannot take the benefit of the bidirectional attention mechanism. However, DSTQA w/o span cannot handle out-of-vocabulary values, but can generalize to new values only by expanding the value sets, moreover, the performance of DSTQA w/o span may decrease when the size of value sets increases.
\begin{wraptable}{r}{0pt}
\resizebox{0.35\textwidth}{!}{%
\begin{tabular}{|l|c|c|}
\hline
                                                       & Joint  & Slot  \\ \hline
TRADE                                                  &   45.60        &   -       \\ \hhline{|=|=|=|}
DSTQA w/span    & \textbf{49.67}  &  97.10    \\ \hline
-graph  &   49.48 &  97.05   \\ \hline
-ELMo +GloVe & 48.15  & 96.98
\\ \hhline{|=|=|=|}
DSTQA w/o span    & \textbf{51.17}  &  97.21    \\ \hline
-ELMo +GloVe  & 50.03  & 97.12     \\ \hline
\end{tabular}%
}
\caption{Results on MultiWOZ $2.1$ dataset.}
\label{exp:woz21}
\end{wraptable}
Table \ref{exp:woz21} shows the results on MultiWOZ $2.1$ dataset. Compared with TRADE, DSTQA w/span has a $8.93\%$ relative improvement and a $4.07\%$ absolute improvement. DSTQA w/o span has a $12.21\%$ relative improvement and a $5.57\%$ absolute improvement. More baselines can be found at the leaderboard.\footnote{http://dialogue.mi.eng.cam.ac.uk/index.php/corpus/} Our model outperforms all models on the leaderboard at the time of submission of this paper.

\textbf{Ablation Study}:
Table \ref{exp:alldomain} also shows the results of ablation study of DSTQA w/span on MultiWOZ $2.0$ dataset. The first experiment completely removes the graph component, and the joint accuracy drops $0.47\%$. The second experiment keeps the graph component but removes the gating mechanism, which is equivalent to setting $\gamma$ in Equation (\ref{eq:graph}) to $0.5$, and the joint accuracy drops $0.98\%$, demonstrating that the gating mechanism is important when injecting graph embeddings and simply adding the graph embeddings to context embeddings can negatively impact the performance. 
In the third experiment, we replace $B_i^{QD}$ with the mean of query word embeddings and replace $B_j^{CD}$ with the mean of context word embeddings. This is equivalent to setting the bi-directional attention scores uniformly. The joint accuracy significantly drops $1.62\%$. The fourth experiment completely removes the bi-directional attention layer, and the joint accuracy drops $1.85\%$. 
Both experiments show that bidirectional attention layer has a notably positive impact on model performance. The fifth experiment substitute ELMo embeddings with GloVe embeddings to demonstrate the benefit of using contextual word embeddings. We plan to try other state-of-the-art contextual word embeddings such as BERT~\citep{devlin2019bert} in the future.  
We further show the model performance on different context lengths in Appendix \ref{sec:contextlength}.

\subsection{Generalization to New Domains}
Table \ref{table:domainexpasion} shows the model performance on new domains. We take one domain in MultiWOZ $2.0$ as the target domain, and the remaining $4$ domains as source domains. Models are trained either from scratch using only $5\%$ or $10\%$ sampled data from the target domain, or first trained on the $4$ source domains and then fine-tuned on the target domain with sampled data. In general, a model that achieves higher accuracy by fine-tuning is more desirable, as it indicates that the model can quickly adapt to new domains given limited data from the new domain. 
In this experiment, we compare DSTQA w/span with TRADE.
As shown in Table \ref{table:domainexpasion}, DSTQA consistently outperforms TRADE when fine-tuning on $5\%$ and $10\%$ new domain data. With $5\%$ new domain data, DSTQA fine-tuning has an average of $43.32\%$ relative improvement over DSTQA training from scratch, while TRADE fine-tuning only has an average of $19.99\%$ relative improvement over TRADE training from scratch. DSTQA w/ graph also demonstrates its benefit over DSTQA w/o graph, especially on the taxi domain. This is because the `taxi' domain is usually mentioned at the latter part of the dialogue, and the destination and departure of the taxi are usually the restaurant, hotel, or attraction mentioned in the previous turns and are embedded in the graph.
\begin{table}[]
\resizebox{1.0\textwidth}{!}{%
\begin{tabular}{|l|c|c|c|c|l|c|c|c|c|l|}
\hline
\multicolumn{1}{|c|}{\begin{tabular}[c]{@{}c@{}}Domain\\ Expansion\end{tabular}} & \multicolumn{2}{c|}{\begin{tabular}[c]{@{}c@{}}Train on 5\% Data\\ from Scratch\end{tabular}} & \multicolumn{3}{c|}{\begin{tabular}[c]{@{}c@{}}Train on 5\% Data\\ by Fine Tuning\end{tabular}}                              & \multicolumn{2}{c|}{\begin{tabular}[c]{@{}c@{}}Train on 10\% Data\\ From Scratch\end{tabular}} & \multicolumn{3}{c|}{\begin{tabular}[c]{@{}c@{}}Train on 10\% Data \\ by Fine Tuning\end{tabular}}                            \\ \hline
                                                                                 & TRADE               & \begin{tabular}[c]{@{}c@{}}DSTQA\\ w/o graph\end{tabular}               & TRADE & \begin{tabular}[c]{@{}c@{}}DSTQA\\ w/o graph\end{tabular} & \begin{tabular}[c]{@{}l@{}}DSTQA\\ w/ graph\end{tabular} & TRADE                & \begin{tabular}[c]{@{}c@{}}DSTQA\\ w/o graph\end{tabular}               & TRADE & \begin{tabular}[c]{@{}c@{}}DSTQA\\ w/o graph\end{tabular} & \begin{tabular}[c]{@{}l@{}}DSTQA\\ w/ graph\end{tabular} \\ \hline
Restaurant                                                                       & 47.31               & 35.33                                                                   & 55.70 & 58.89                                                     & \textbf{58.95}                                                    & 53.65                & 54.27                                                                   & 60.94 & \textbf{64.51}                                                     & 64.48                                                    \\ \hline
Hotel                                                                            & 31.93               & 33.08                                                                   & 37.45 & 48.94                                                     & \textbf{50.18}                                                    & 41.29                & 49.69                                                                   & 41.42 & 52.59                                                     & \textbf{53.68}                                                    \\ \hline
Train                                                                            & 48.82               & 50.36                                                                   & 69.27 & 69.32                                                     & \textbf{70.35}                                                    & 59.65                & 61.28                                                                   & 71.11 & 73.74                                                     & \textbf{74.50}                                                    \\ \hline
Attraction                                                                       & 52.19               & 51.58                                                                   & 57.55 & \textbf{70.47}                                                     & 70.10                                                    & 58.46                & 61.77                                                                   & 63.12 & \textbf{71.60}                                                     & 71.28                                                    \\ \hline
Taxi                                                                             & 59.03               & 58.25                                                                   & 66.58 & 68.19                                                     & \textbf{70.90}                                                    & 60.51                & 59.35                                                                   & 70.19 & 72.52                                                     & \textbf{74.19}                                                    \\ \hline
\end{tabular}%
}
\caption{Joint accuracy on domain expansion experiments. Models are either trained from scratch on the target domain, or trained from the $4$ source domains and then fine-tuned on the target domain.}
\label{table:domainexpasion}
\vspace{-0.7cm}
\end{table}

\subsection{Error Analysis}
\begin{wrapfigure}{r}{0.5\textwidth}   \includegraphics[width=0.5\textwidth]{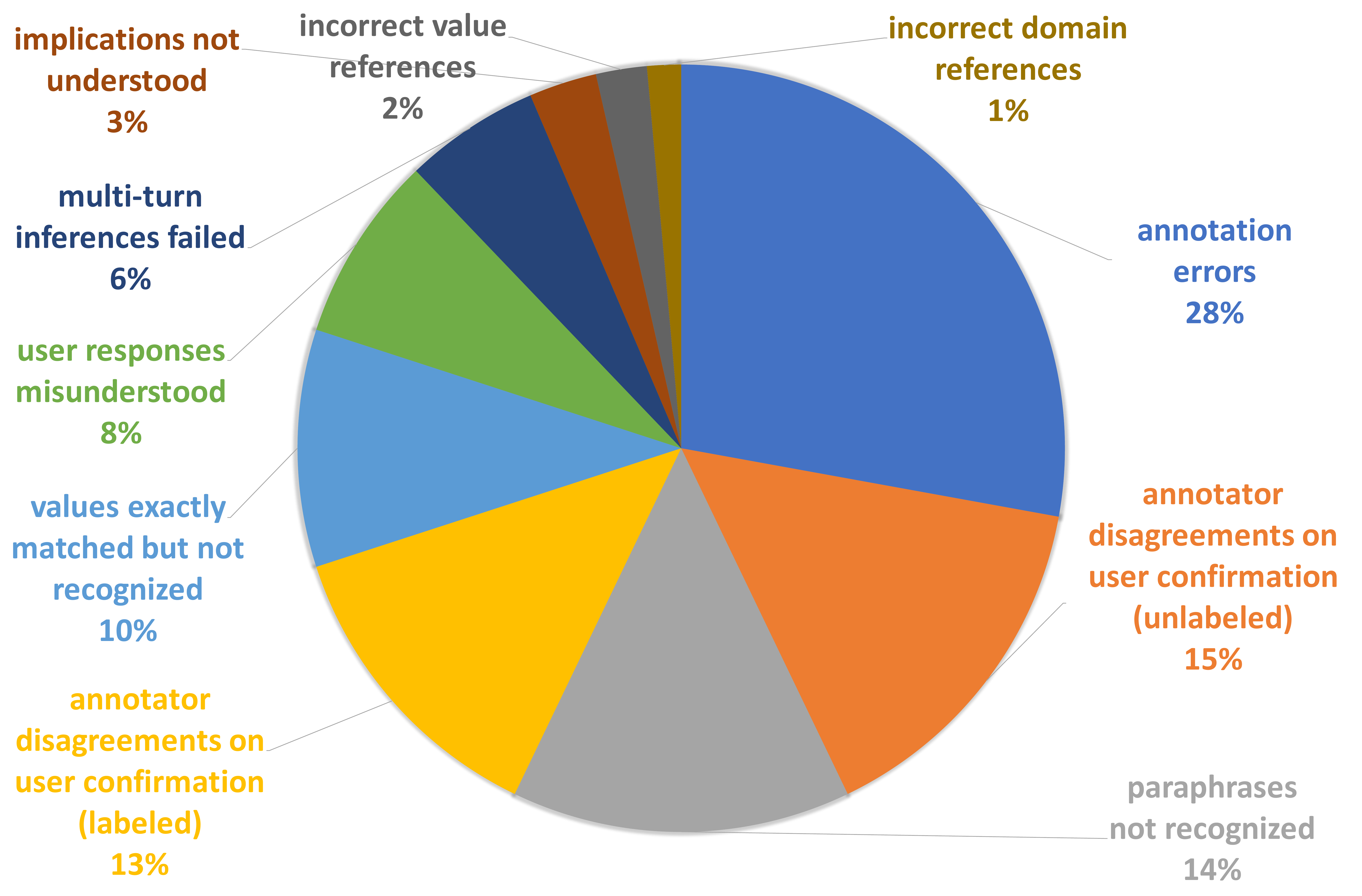}
  \caption{Error Types on MultiWOZ 2.1 dataset.}
  \label{fig:erroranalysis}
\end{wrapfigure}
Figure \ref{fig:erroranalysis} shows the different types of model prediction errors on MultiWOZ $2.1$ dataset made by DSTQA w/span as analyzed by the authors. Appendix \ref{sec:error_examples} explains the meaning of each error type and also list examples for each error type.
At first glance, annotation errors and annotation disagreements account for $56\%$ of total prediction errors, and are all due to noise in the dataset and thus unavoidable. \textit{Annotation errors} are the most frequent errors and account for $28\%$ of total prediction errors. Annotation errors means that the model predictions are incorrect only because the corresponding ground truth labels in the dataset are wrong. Usually this happens when the annotators neglect the value informed by the user. \textit{Annotator disagreement on user confirmation} accounts for $28\%$ ($15\% + 13\%$) of total errors. This type of errors comes from the disagreement between annotators when generating ground truth labels. 
All these errors are due to the noise in the dataset and unavoidable, which also explains why the task on MultiWOZ $2.1$ dataset is challenging and the state-of-the-art joint accuracy is less than $50\%$.

\textit{Values exactly matched but not recognized} ($10\%$) and \textit{paraphrases not recognized} ($14\%$) mean that the user mentions a value or a paraphrase of a value, but the model fails to recognize it. \textit{Multi-turn inferences failed} ($6\%$) means that the model fails to refer to previous utterances when making prediction. \textit{User responses not understood} ($8\%$) and \textit{implications not understood} ($3\%$) mean that the model does not understand what the user says and fails to predict based on user responses. Finally, \textit{incorrect value references} ($2\%$) means that there are multiple values of a slot in the context and the model refers to an incorrect one, and \textit{incorrect domain references} ($1\%$) means that the predicted slot and value should belong to another domain. All these errors indicate insufficient understanding of agent and user utterances. A more powerful language model and a coreference resolution modules may help mitigate these problems. Please refer to Appendix \ref{sec:error_examples} for examples.

\section{Related Works}
\label{sec:related_works}
Our work is most closely related to previous works in dialogue state tracking and question answering. Early models of dialogue state tracking~\citep{thomson2010bayesian,wang2013simple,henderson2014robust} rely on handcrafted features to extract utterance semantics, and then use these features to predict dialogue states. Recently~\citet{mrkvsic2017neural} propose to use convolutional neural network to learn utterance $n$-gram representation, and achieve better performance than handcrafted features-based model. However, their model maintains a separate set of parameters for each slot and does not scale well. Models that handles scalable multi-domain DST have then been proposed~\citep{ramadan2018large,rastogi2017scalable}. ~\citet{zhong2018global} and~\citet{gce} propose a global-local architecture. The global module is shared by all slots to transfer knowledge between them.~\citet{ren2018towards} propose to share all parameters between slots and fix the word embeddings during training, so that they can handle new slots and values during inference. However, These models do not scale when the sizes of value sets are large or infinite, because they have to evaluate every (domain, slot, tuple) during the training.~\citet{xu2018end} propose to use a pointer network with a Seq2Seq architecture to handle unseen slot values.~\citet{lee2019sumbt} encode slots and utterances with a pre-trained BERT model, and then use a slot utterance matching module, which is a multi-head attention layer, to compute the similarity between slot values and utterances.~\citet{rastogi2019towards} release a schema-guided DST dataset which contains natural language description of domains and slots. They also propose to use BERT to encode these natural language description as embeddings of domains and slots.
~\citet{wu-etal-2019-transferable} propose to use an encoder-decoder architecture with a pointer network. The source sentences are dialogue contexts and the target sentences are annotated value labels. The model shares parameters across domains and does not require pre-defined domain ontology, so it can adapt to unseen domains, slots and values. Our work differs in that we formulate multi-domain DST as a question answering problem and use reading comprehension methods to provide answers. There have already been a few recent works focusing on using reading comprehension models for dialogue state tracking. For example,~\citet{perez2017dialog} formulate slot tracking as four different types of questions (Factoid, Yes/No, Indefinite knowledge, Counting and Lists/Sets), and use memory network to do reasoning and to predict answers.~\citet{gao2019dialog} construct a question for each slot, which basically asks {\em what is the value of slot i}, then they predict the span of the value/answer in the dialogue history. Our model is different from these two models in question representation. We not only use domains and slots but also use lists of candidate values to construct questions. Values can be viewed as descriptions to domains and slots, so that the questions we formulate have richer information about domains and slots, and can better generalize to new domains, slots, and values. Moreover, our model can do both span and value prediction, depending on whether the corresponding value lists exists or not. Finally, our model uses a dynamically-involving knowledge graph to explicitly capture interactions between domains and slots.

In a reading comprehension~\citep{rajpurkar2016squad} task, there is one or more context paragraphs and a set of questions. The task is to answer questions based on the context paragraphs. Usually, an answer is a text span in a context paragraph. Many reading comprehension models have been proposed~\citep{seo2016bidirectional,yu2018qanet,devlin2019bert,clark2018simple,chen2017reading}. These models encode questions and contexts with multiple layers of attention-based blocks and predict answer spans based on the learned question and context embeddings.
Some works also explore to further improve model performance by knowledge graph. For example~\citet{sun2018open} propose to build a heterogeneous graph in which the nodes are knowledge base entities and context paragraphs, and nodes are linked by entity relationships and entity mentions in the contexts.~\citet{zhang2018kg} propose to use Open IE to extract relation triples from context paragraphs and build a contextual knowledge graph with respect to the question and context paragraphs. We would expect many of these technical innovations to apply given our QA-based formulation.

\section{Conclusion}
In this paper, we model multi-domain dialogue state tracking as question answering with a dynamically-evolving knowledge graph. 
Such formulation enables the model to generalize to new domains, slots and values by simply constructing new questions. Our model achieves state-of-the-art results on MultiWOZ 2.0 and MultiWOZ 2.1 dataset with a $5.80\%$ and a $12.21\%$ relative improvement, respectively. Also, our domain expansion experiments show that our model can better adapt to unseen domains, slots and values compared with the previous state-of-the-art model.

\bibliography{iclr2020_conference}
\bibliographystyle{iclr2020_conference}

\newpage
\appendix
\section{Appendix}
\subsection{Results on WOZ 2.0 dataset}
\label{ap:woz2}
We also evaluate our algorithm on WOZ $2.0$
dataset~\citep{mrkvsic2017neural}
\begin{wraptable}{r}{0pt}
\begin{tabular}{|l|l|}
\hline
Model        & Joint Accuracy     \\ \hline
NBT          & 84.4          \\ \hline
GLAD         & 88.1          \\ \hline
GCE         & 88.5          \\ \hline
StateNet PSI & 88.9          \\ \hline
SUMBT & \textbf{91.00} \\ \hline
DSTQA        & 90.0 \\ \hline
\end{tabular}%
\caption{Joint accuracy on WOZ 2.0 dataset.}
\label{exp:woz2}
\end{wraptable}
WOZ $2.0$ dataset has $1200$ restaurant domain task-oriented dialogues. There are three slots: `food', `area', `price range', and a total of $91$ slot values. The dialogues are collected from a Wizard of Oz style experiment, in which the task is to find a restaurant that matches the slot values the user has specified. Each turn of a dialogue is annotated with a dialogue state, which indicates the slot values the user has informed. One example of the dialogue state is \textit{\{`food:Mexican', `area':`east', price range:`moderate'\}}.

Table \ref{exp:woz2} shows the results on WOZ $2.0$ dataset. We compare with four published baselines. SUMBT~\citep{lee2019sumbt} is the current state-of-the-art model on WOZ 2.0 dataset. It fine-tunes a pre-trained BERT model~\citep{devlin2019bert} to learn slot and utterance representations. StateNet PSI~\citep{ren2018towards} maps contextualized slot embeddings and value embeddings into the same vector space, and calculate the Euclidean distance between these two. It also learns a joint model of all slots, enabling parameter sharing between slots. GLAD~\citep{zhong2018global} proposes to use a global module to share parameters between slots and a local module to learn slot-specific features. Neural Beflief Tracker~\citep{mrkvsic2017neural} applies CNN to learn n-gram utterance representations. Unlike prior works that transfer knowledge between slots by sharing parameters, our model implicitly transfers knowledge by formulating each slot as a question and learning to answer all the questions. Our model has a $1.24\%$ relative joint accuracy improvement over StateNet PSI. Although SUMBT achieves higher joint accuracy than DSTQA on WOZ $2.0$ dataset, DSTQA achieves better performance than SUMBT on MultiWOZ $2.0$ dataset, which is a more challenging dataset.

\subsection{MultiWOZ 2.0/2.1 Ontology}
\label{ap:ontology}
The ontology of MultiWOZ $2.0$ and MultiWOZ $2.1$ datasets is shown in Table \ref{table:ontology}. There are $5$ domains and $30$ slots in total. (two other domains `hospital' and `police' are ignored as they only exists in training set.)
\begin{table}[h]
\centering
\begin{tabular}{|c|c|c|c|c|c|}
\hline
Domains & Restaurant                                                                                                    & Hotel                                                                                                                                             & Train                                                                                                      & Attraction                                                 & Taxi                                                                                   \\ \hline
Slots   & \begin{tabular}[c]{@{}c@{}}name\\ area\\ price range\\ food\\ book people\\ book time\\ book day\end{tabular} & \begin{tabular}[c]{@{}c@{}}name\\ area\\ price range\\ type\\ parking\\ stars\\ internet\\ book stay\\ book day\\ book people\end{tabular} & \begin{tabular}[c]{@{}c@{}}destination\\ departure\\ day\\ arrive by\\ leave at\\ book people\end{tabular} & \begin{tabular}[c]{@{}c@{}}name\\ area\\ type\end{tabular} & \begin{tabular}[c]{@{}c@{}}destination\\ departure\\ arrive by\\ leave at\end{tabular} \\ \hline
\end{tabular}
\caption{Domain ontology in MultiWOZ 2.0 and MultiWOZ 2.1 dataset}
\label{table:ontology}
\end{table}

\subsection{Performance on Each Individual Domain}
\label{ap:singledomain}
\begin{table}[h]
\centering
\begin{tabular}{|l|c|c|c|c|}
\hline
           & \multicolumn{2}{c|}{Joint Accuray} & \multicolumn{2}{c|}{Slot Accuracy} \\ \hline
           & TRADE            & DSTQA w/span           & TRADE            & DSTQA w/span          \\ \hline
Restaurant & 65.35            & \textbf{68.68}           & 93.28            & 94.08           \\ \hline
Hotel      & 55.52            & \textbf{61.76}           & 92.66            & 93.72           \\ \hline
Train      & 77.71            & \textbf{79.75}           & 95.30            & 95.61           \\ \hline
Attraction & 71.64            & \textbf{74.05}           & 88.97            & 90.53           \\ \hline
Taxi       & 76.13            & \textbf{78.22}           & 89.53            & 90.37           \\ \hline
\end{tabular}%
\caption{Model performance on each of the $5$ domains.}
\label{table:sd}
\end{table}

We show the performance of DSTQA w/span and TRADE on each single domain. We follow the same procedure as~\citet{wu-etal-2019-transferable} to construct training and test dataset for each domain: a dialogue is excluded from a domain's training and test datasets if it does not mention any slots from that domain. During the training, slots from other domains are ignored. Table \ref{table:sd} shows the results. We can see that our model achieves better results on every domain, especially the hotel domain, which has a $11.24\%$ relative improvement. Hotel is the hardest domain as it has the most slots (10 slots) and has the lowest joint accuracy among all domains.

\begin{wrapfigure}{r}{0.4\textwidth}  \includegraphics[width=0.4\textwidth]{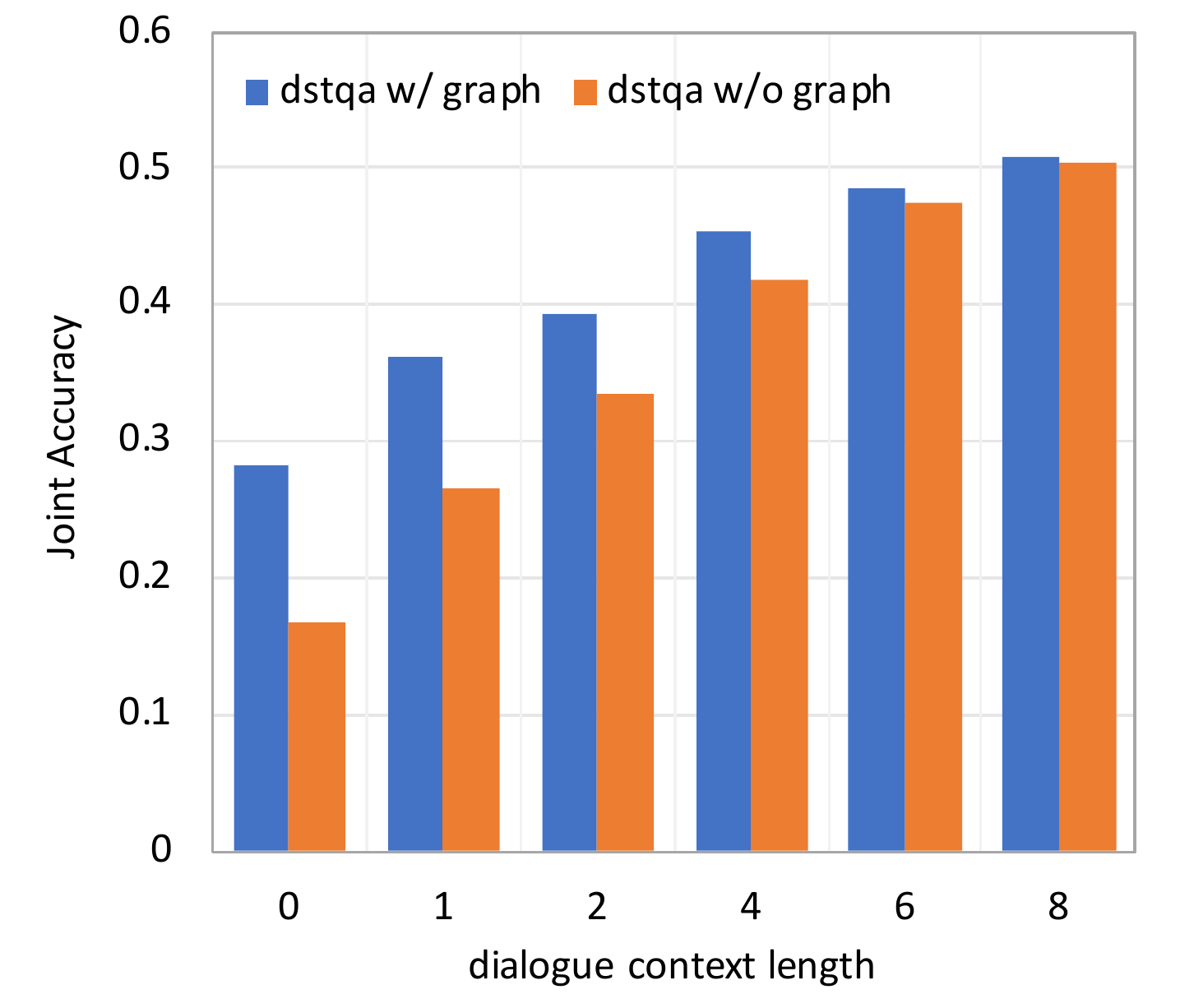}
  \caption{Joint acc. v.s. context length}
  \label{fig:contextlength}
  \vspace{-0.5cm}
\end{wrapfigure}

\subsection{Joint Accuracy v.s. Context Length}
\label{sec:contextlength}
We further show the model performance on different context lengths. Context lengths means the number of previous turns included in the dialogue context. Note that our baseline algorithms either use all previous turns as contexts to predict belief states or accumulate turn-level states of all previous turns to generate belief states. The results are shown in Figure \ref{fig:contextlength}. We can see that DSTQA with graph outperforms DSTQA without graph. This is especially true when the context length is short. This is because when the context length is short, graph carries information over multiple turns which can be used for multi-turn inference. This is especially useful when we want a shorter context length to reduce computational cost. In this experiment, the DSTQA model we use is DSTQA w/span.

\subsection{Accuracy per Slot}
\begin{figure}[h]
\hspace{-1.3cm}
  \includegraphics[width=1.15\textwidth]{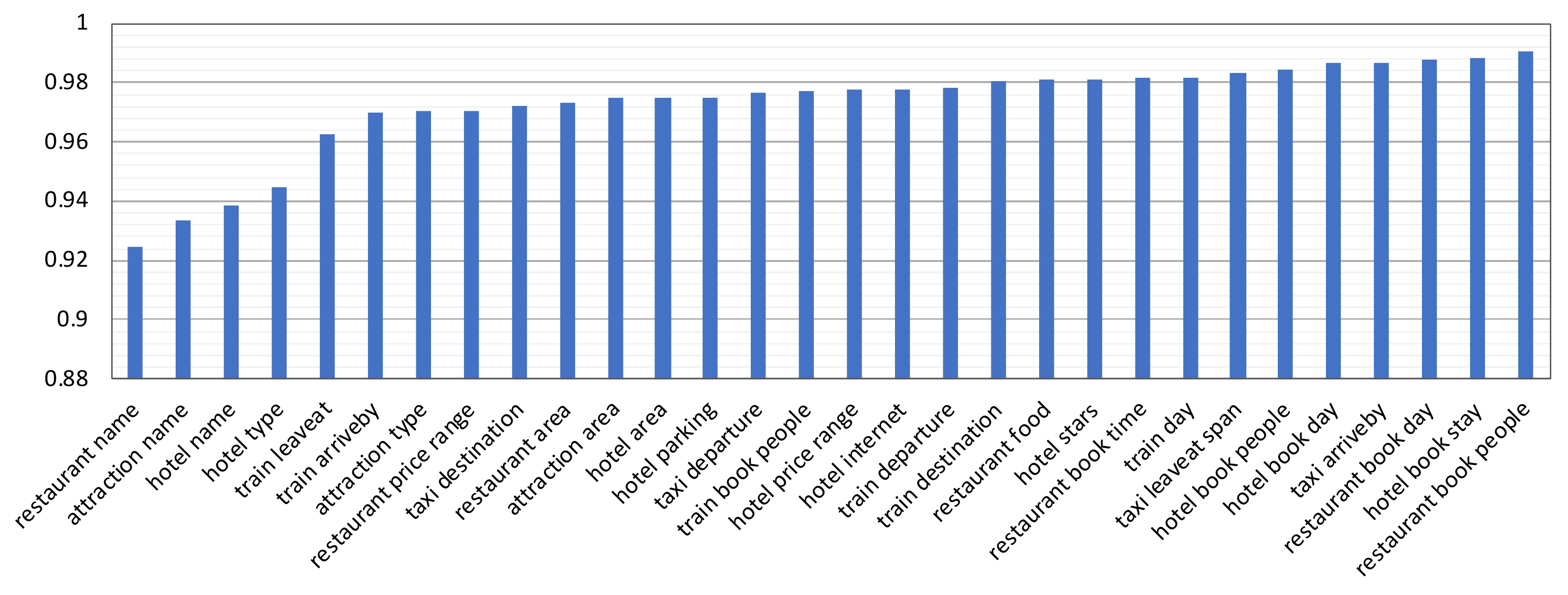}
  \caption{Accuracy of each slot per turn on MultiWOZ 2.0 dataset}
  \label{fig:erroranalysis1}
\end{figure}
\begin{figure}[h]
\hspace{-1.3cm}
  \includegraphics[width=1.15\textwidth]{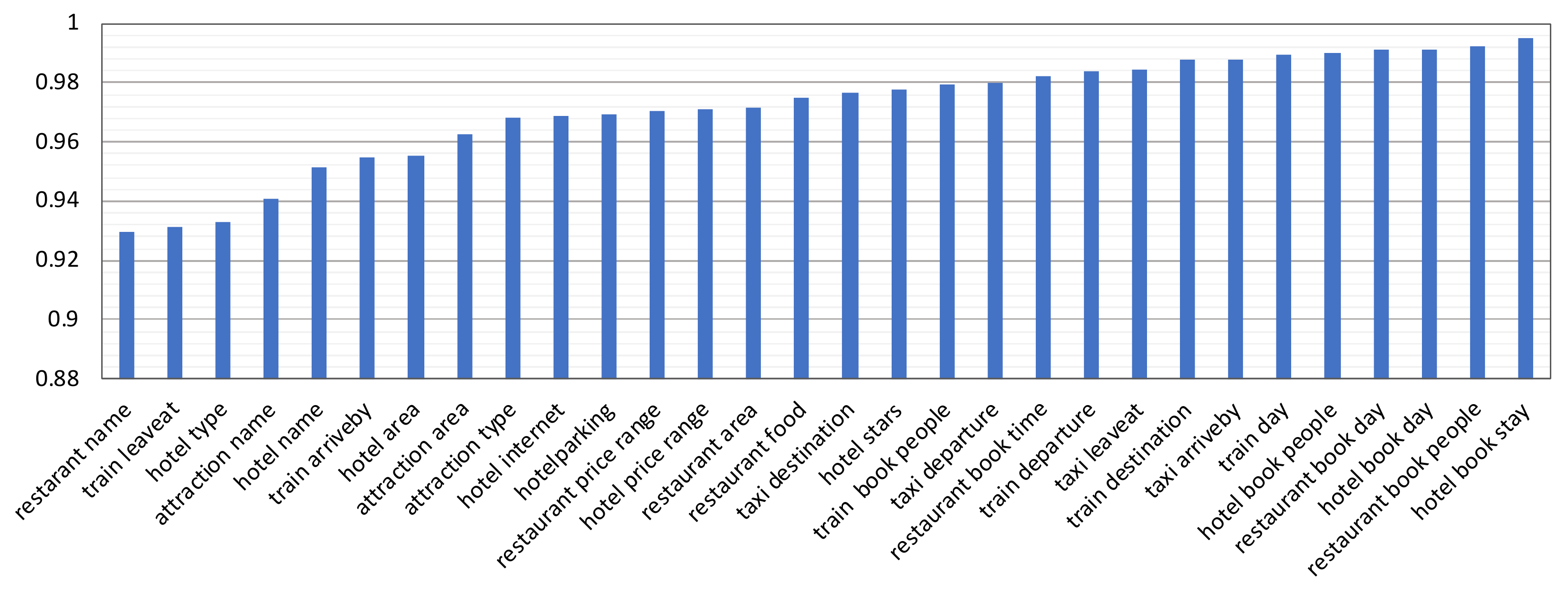}
  \caption{Accuracy of each slot per turn on MultiWOZ 2.1 dataset}
  \label{fig:erroranalysis2}
\end{figure}
The accuracy of each slot on MultiWOZ $2.0$ and MultiWOZ $2.1$ test set is shown in Figure \ref{fig:erroranalysis1} and Figure \ref{fig:erroranalysis2}, respectively. Named related slots such as {\em restaurant name}, {\em attraction name}, {\em hotel name} has high error rate, because these slots have very large value set and high annotation errors. 

\subsection{Examples of Prediction Errors}
\label{sec:error_examples}
This section describes prediciton errors made by DSTQA w/span. Incorrectly predicted (domain, slot, value) tuples are marked by underlines.
\subsubsubsection{\textbf{1.\ Annotation errors}} \\
Description: The groud truth label in the dataset is wrong. This can happen either 1) annotators neglect slots mentioned in the user utterance 2) annotators mistakenly choose the wrong label of a slot. \\
Examples: 

\fbox{\parbox{\textwidth}{
User: I would like to find a \textit{museum} in the \textit{west} to go to. \\
Agent: There are several museums in the west. I recommend the \textit{Cafe Jello Gallery}. \\
User: Can I have the address of the Cafe Jello museum? \\
Agent: The Cafe Jello Gallery is at 13 Magdalene street. Is there anything else? \\
User: Is there a \textit{moderately} priced \textit{British} restaurant \textit{any where} in town? 
}
}
\fbox{\parbox{\textwidth}{
\textbf{Annotation}: \{(restaurant, food, British), (restaurant, price range, moderate), \ul{(restaurant, area, west)}\} \\
\textbf{Prediction}: \{(restaurant, food, Biritsh), (restaurant, price range, moderate), \ul{(restaurant, area, don't care)}\}
}
}

\subsubsubsection{\textbf{2.\ Annotator disagreement on user confirmation (labeled)}} \\
Description: 
This type of errors comes from the disagreement between annotators when generating ground truth labels. More specifically, in a dialogue, the agent sometimes proposes a suggestion (a value of a slot) to the user, followed by the user's positive confirmation. For example, the agent says `I would recommend Little Seoul. Would you like to make a reservation?'. The user confirms with `yes, please'. Since the user positively confirms the agent’s suggestion, the (domain, slot, value) tuple mentioned by the agent, or, (restaurant, name, Little Seoul) tuple in this example, can be added into the belief state. However, based on our observation of the MultiWOZ 2.0 and MultiWOZ 2.1 dataset, the annotators are inconsistent, and only about half of the times these tuples are added to the belief states. An error of this type comes from the scenario that the tuple is added to the belief state by the annotator but not by the model (i.e. the model predicts ‘None’ for the corresponding (domain, slot) pair). \\
Examples:

\fbox{\parbox{\textwidth}{
User: I would like to find an \textit{architectural attraction} to go to in the \textit{centre}. Can you help me? \\
Agent: Yes! I found 5 attractions matching your requirements . Would you like to know more?  \\
User: I would like to get the entrance fee and address of them please. \\
Agent: \textit{Great Saint Mary's Church} is 2 pounds and the other 4 are free. Would you like the address for Great Saint Mary's church? \\
User: \textit{Yes}, that would be great. 
}
}
\fbox{\parbox{\textwidth}{
\textbf{Annotation}: \{(attraction, type, architecture), (attraction, area, centre), \ul{(attraction, name, Great Saint Mary's Church)}\} \\
\textbf{Prediction}: \{(attraction, type, architecture), (attraction, area, centre)\}  
}
}
\subsubsubsection{\textbf{3.\ Annotator disagreement on user confirmation (unlabeled)}} \\
Description: Similar to the last type of errors, this one comes from the scenario that the tuple is added to the belief state by the model but not by the annotator.  \\
Examples:

\fbox{\parbox{\textwidth}{
Agent: There are many great restaurants in the \textit{centre}. Is there a certain type of food you would like to have? \\
User: What do you have that is \textit{expensive}? \\
Agent: Yes, there are many. Would you be interested in trying British cuisine or African? \\
User: \textit{British cuisine} is fine. Can you suggest a restaurant? \\
Agent: \textit{Fitzbillies restaurant} meets all of your requirements. Would you like more info on them? \\
User: Book this for \textit{6 people} on \textit{Monday} at \textit{12:30}.
}
}
\fbox{\parbox{\textwidth}{
\textbf{Annotation}: \{(restaurant, area, centre), (restaurant, food, British), (restaurant, price range, expensive), (restaurant, time, 12:30), (restaurant, book people, 6), (restaurant, day, Monday)\} \\
\textbf{Prediction}: \{(restaurant, area, centre), (restaurant, food, British), (restaurant, price range, expensive), (restaurant, time, 12:30), (restaurant, book people, 6), (restaurant, day, Monday), \ul{(restaurant, name, Fitzbillies restaurant)}\}  
}
}

\subsubsubsection{\textbf{4.\ Paraphrases not recognized}} \\
Description: The paraphrases of a value is not recognized by the model. \\
Example:

\fbox{\parbox{\textwidth}{
User: Can you help me find a place to go in the centre? \\
Agent: I can help you with that. Is there a certain kind of attraction that you would like to visit? \\
User: Surprise me. Give me the postcode as well. 
}
}
\fbox{\parbox{\textwidth}{
\textbf{Annotation}: \{(attraction, area, centre), \ul{(attraction, area, don't care)}\} \\
\textbf{Prediction}: \{(attraction, area, centre)\}  
}
}

\subsubsubsection{\textbf{5.\ Value exactly matched but not recognized}} \\
Description:
The value of a slot is mentioned and exactly matched in the user's utterance, but the model fails recognize and predict it. \\
Examples:

\fbox{\parbox{\textwidth}{
Agent: I am sorry, there is no restaurant serving specifically North American or American food in my database, is there another type of food you would consider? \\
User: How about \textit{Modern European} food? \\
Agent: There are 3 Modern European restaurants. Two in the centre and one in the south. Do you have a preference? \\
User: I would prefer the one on the \textit{centre}, could I have the phone number and postcode please?
}
}
\fbox{\parbox{\textwidth}{
\textbf{Annotation}: \{(restaurant, food, Modern European), \ul{(restaurant, area, centre)}\} \\
\textbf{Prediction}: \{(restaurant, food, Modern European)\}  
}
}

\subsubsubsection{\textbf{6.\ User responses misunderstood}} \\
Description: The model misunderstands the user's intention and fails to predict based on the user utterance. \\
Examples:

\fbox{\parbox{\textwidth}{
User: I could use some help finding a restaurant that is moderately priced. \\
Agent: We have many options that are \textit{moderately priced}. Is there a specific area or type of cuisine you are looking for? \\
User: I do not care about the cuisine but I want it to be \textit{in the west}. \\
Agent: We have \textit{Prezzo}. It is an Italian restaurant located in the west. it is moderately priced. Would you like me to book it for you? \\
User: \textit{That will not be necessary}. What is the postcode? \\
Agent: Prezzo's postcode is cb30ad.
}
}
\fbox{\parbox{\textwidth}{
\textbf{Annotation}: \{(restaurant, price range, moderate), (restaurant, area, west)\} \\
\textbf{Prediction}: \{(restaurant, price range, moderate), (restaurant, area, west), \ul{(restaurant, name, Prezzo)}\}  
}
}

\subsubsubsection{\textbf{7.\ Multi-turn inference failed}} \\
Description: In this scenario, it requires information from multiple turns to predict the value of a slot, but the model fails to perform multi-turn inference. \\
Example:

\fbox{\parbox{\textwidth}{
User: Hello, may I have a list of museums in the west? \\
Agent: There are 7: Cafe Jello Gallery, Cambridge and County Folk Museum, ... \\
User: Please give me the entrance fee and postcode of County Folk Museum \\
Agent: The entrance fee is $3.50$ pounds and the postcode is cb30aq. Would you like any other information? \\
User: I need a place to eat \textit{near the museum}. I do not want to spend much so it should be \textit{cheap}. what do you have?
}
}
\fbox{\parbox{\textwidth}{
\textbf{Annotation}: \{(attraction, area, west), (attraction, type, museum), (attraction, name, Cambridge and County Folk Museum), (restaurant, price range, cheap), \ul{(restaurant, area, centre)}\} \\
\textbf{Prediction}: \{(attraction, area, west), (attraction, type, museum), (attraction, name, Cambridge and County Folk Museum), (restaurant, price range, cheap)\}  
}
}

\subsubsubsection{\textbf{8.\ Implication not understood}} \\
Description: Implication expressed by the user is not understood by the model. \\
Examples: 

\fbox{\parbox{\textwidth}{
User: I am trying to find a train leaving after \textit{14:45} that's heading out \textit{from London Liverpool street}. What do you have? \\
Agent: There are 45 trains that fit your criteria. Please clarify your destination, day of travel and the time you want to arrive by so that i can narrow it down. \\
User: I need a train \textit{to Cambridge} on Tuesday. \\
Agent: I have 5 departures fitting your criteria on the :39 of the hour from 15:39 to 23:39. Would you like me to book any of these for you ? \\
User: Yes please do book the 15:39.
}
}
\fbox{\parbox{\textwidth}{
\textbf{Annotation}: \{(train, leaveat, 14:45), (train, departure, London Liverpool street), (train, destination, Cambridge), (train, day, Tuesday), \ul{(train, book people, 1)}\} \\
\textbf{Prediction}: \{(train, leaveat, 14:45), (train, departure, London Liverpool street), (train, destination, Cambridge), (train, day, Tuesday)\}  
}
}

\subsubsubsection{\textbf{9.\ Incorrect value reference}} \\
Description: There are multiple values of a slot in the context and the model refers to an incorrect one. This usually happens in time-related slots such as train departure time. \\
Examples:

\fbox{\parbox{\textwidth}{
User: I need to travel on \textit{Saturday} from \textit{Cambridge} to \textit{London Kings Cross} and need to leave after \textit{18:30}. \\
Agent: Train tr0427 leaves at 19:00 on Saturday and will get you there by 19:51. the cost is 18.88 pounds. Want me to book it? \\
User: Yes, please book the train for \textit{1 person} and provide the reference number.
}
}
\fbox{\parbox{\textwidth}{
\textbf{Annotation}: \{(train, departure, Cambridge), (train, destination, London King Cross), (train, day, Saturday), (train, book people, 1), \ul{(train, leaveat, 18:30)}\} \\
\textbf{Prediction}: \{(train, departure, Cambridge), (train, destination, London King Cross), (train, day, Saturday), (train, book people, 1), \ul{(train, leaveat, 19:00)}\}  
}
}

\subsubsubsection{\textbf{10.\ Incorrect domain reference}} \\
Description: The predicted slot and value should belong to another domain. This happens because many slots exists in multiple domains.  \\
Example:

\fbox{\parbox{\textwidth}{
User: I am looking for information on Cambridge University Botanic Gardens. \\
Agent: They are on Bateman st., postal code cb21jf. They can be reached at 01223336265, the entrance fee is 4 pounds. Can I help with anything else? \\
User: Yes, can you help me find a restaurant? \\
Agent: The botanic gardens are in the centre . Would you like the restaurant to also be in the centre? do you have any type of cuisine in mind? \\
User: \textit{never mind, i will worry about food later}. I am actually looking for a hotel with a \textit{guesthouse} and \textit{free parking} would be great as well. \\
Agent: There are 21 guesthouses with free parking, do you have a price or area preference? \\
User: \textit{cheap} and \textit{in the south} please .
}
}
\fbox{\parbox{\textwidth}{
\textbf{Annotation}: \{(hotel, area, south), (hotel, parking, yes), (hotel, price range, cheap), (hotel, type, guesthouse)\} \\
\textbf{Prediction}: \{(hotel, area, south), (hotel, parking, yes), (hotel, price range, cheap), (hotel, type, guesthouse), \ul{(restaurant, price range, cheap)}\}  
}
}

\end{document}